\documentclass[letterpaper]{article} 
\usepackage{aaai24}  
\usepackage{times}  
\usepackage{helvet}  
\usepackage{courier}  
\usepackage[hyphens]{url}  
\usepackage{graphicx} 
\usepackage{natbib}  
\usepackage{caption} 
\usepackage{algorithm}
\usepackage{algorithmic}

\usepackage{newfloat}
\usepackage{listings}
\DeclareCaptionStyle{ruled}{labelfont=normalfont,labelsep=colon,strut=off} 
\floatstyle{ruled}
\newfloat{listing}{tb}{lst}{}
\floatname{listing}{Listing}
\title{LaViP: Language-Grounded Visual Prompts}
\author {
    Nilakshan Kunananthaseelan\textsuperscript{\rm 1},
    Jing Zhang\textsuperscript{\rm 2},
    Mehrtash Harandi\textsuperscript{\rm 1}
}
\affiliations {
    \textsuperscript{\rm 1}Department of Electrical and Computer Systems Engineering, Monash University\\
    \textsuperscript{\rm 2}College of Engineering and Computer Science, Australian National University
\\
    \texttt{\{nilakshan.kunananthaseelan,mehrtash.harandi\}@monash.edu, jing.zhang@anu.edu.au}
}

\usepackage{bibentry}
\usepackage{booktabs}
\usepackage{import}
\usepackage{amsmath}
\usepackage{csquotes}
\usepackage{subcaption}
\usepackage{colortbl}
\usepackage{multirow}
\usepackage[multiple]{footmisc}
\usepackage{amsmath}
\usepackage{mathrsfs}
\usepackage{amsfonts}
\usepackage{amssymb}
\usepackage{mathtools}
\usepackage[table,dvipsnames]{xcolor}
\usepackage{graphicx, xspace}

\usepackage{url}

\usepackage[pagebackref,breaklinks,colorlinks,citecolor=black]{hyperref}
\usepackage{cleveref}
\definecolor{my_purple}{HTML}{9903F0}
\definecolor{my_green}{HTML}{156C09}
\definecolor{my_orange}{HTML}{FF3300}

\DeclareGraphicsExtensions{.pdf,.jpeg,.png,.jpg,.eps}

\makeatletter

\DeclareRobustCommand\onedot{\futurelet\@let@token\@onedot}
\def\@onedot{\ifx\@let@token.\else.\null\fi\xspace}

\def\eg{\emph{e.g}\onedot}

\def\etal{\emph{et al}\onedot}
\makeatother

\usepackage{bm}

\def\vnu{{\bm{\nu}}}

\def\vx{{\bm{x}}}
\def\vy{{\bm{y}}}

\def\mA{{\bm{A}}}
\def\mB{{\bm{B}}}

\def\mM{{\bm{M}}}

\def\mX{{\bm{X}}}
\def\mY{{\bm{Y}}}

\begin{document}

\maketitle


\begin{abstract}
  We introduce a language-grounded visual prompting method to adapt the visual encoder of vision-language models for downstream tasks. By capitalizing on language integration, we devise a parameter-efficient strategy to adjust the input of the visual encoder, eliminating the need to modify or add to the model's parameters. Due to this design choice, our algorithm can operate even in black-box scenarios, showcasing adaptability in situations where access to the model's parameters is constrained. We will empirically demonstrate that, compared to prior art, grounding visual prompts with language enhances both the accuracy and speed of adaptation. Moreover, our algorithm excels in base-to-novel class generalization, overcoming limitations of visual prompting and exhibiting the capacity to generalize beyond seen classes. We thoroughly assess and evaluate our method across a variety of image recognition datasets, such as EuroSAT, UCF101, DTD, and CLEVR, spanning different learning situations, including few-shot learning, base-to-novel class generalization, and transfer learning.

\end{abstract}

\section{Introduction}
Large-scale pre-trained models (PTMs)~\cite{ref_gpt,ref_vit,ref_clip,ref_llama,ref_sam} are trained on massive amounts of data and intricate optimization throughout the learning process.
This makes designing and developing high-performing PTMs a laborious and costly process. 
While these models showcase generalization prowess, achieving optimal performance on new tasks necessitates careful finetuning.
Nonetheless, the finetuning of PTMs carries inherent challenges, notably the risk of catastrophic knowledge forgetting and vulnerability to overfitting on the downstream tasks~\cite{kumar2021fine,ref_finetuning2}.
\begin{figure}[!t]
\centering
\includegraphics[width=1.1\columnwidth]{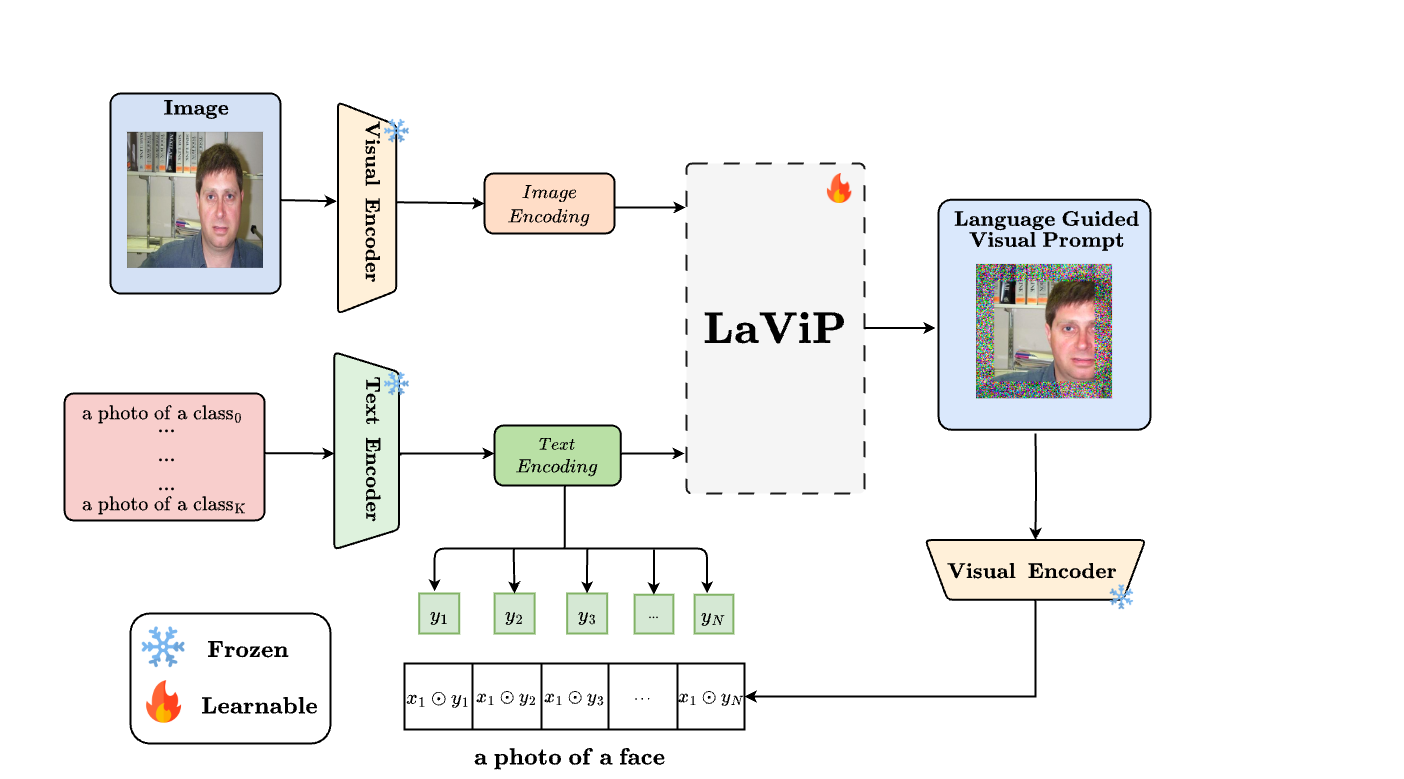} 
\caption{Our key idea is to reprogram the visual encoder of CLIP~\cite{ref_clip} through the generation of language-grounded visual prompts.}
\label{fig:concept}
\end{figure}

In response to the aforementioned challenges, \textit{model reprogramming(MR)}~\cite{ref_ar}, a method in the context of transfer learning, has emerged as a fresh paradigm. The core idea behind MR is to repurpose and harness a high-quality pre-trained model, facilitating seamless cross-domain learning \textit{without the need for finetuning} the model. MR introduces a learnable transformation function at the input of the model, along with an output mapping function to achieve this objective. The pioneering work of~\cite{ref_bar} has demonstrated that through MR, even a CNN initially trained on ImageNet can be swiftly adapted to excel in classifying medical images, interestingly, even outperforming the traditional finetuning approach. Subsequent research efforts have extended the idea of MR into various domains, achieving successful adaption without finetuing~\cite{ref_mr1,ref_mr2,ref_mr3,ref_mr4,ref_lblmap}.

The input transformation acquired through MR is commonly conceptualized as a perturbation pattern, which is either added to or concatenated with the input images. By learning the perturbation pattern, also called Visual Prompts (VPs) in vision tasks, the PTM effectively embeds the downstream task samples into a distinct subset of its latent space. As such, MR allows to adeptly repurpose the PTM's capabilities, all while preserving the integrity of the latent space.  Despite the promise and rapid progress, several questions remain unanswered in MR;
\begin{itemize}
     \item \textbf{Unimodality in learning VPs.}  
     To the best of our knowledge, in the previous studies focusing on VPs, class semantic information and visual encoding are typically treated separately in many cases, despite human perception being multimodal (\eg,~\cite{ref_neural1, ref_neural2, ref_neural3}. This multimodal framework of our cognitive system helps us to learn new concepts with a few examples.
     This, in AI, will raise a simple question, 
     \textit{if a Vision-Language model is at hand, does language help in designing VPs for MR? If yes, what are the design questions to answer?}
     \item \textbf{Efficient Training.} In practice, learning VPs require a large number of iterations to achieve quality results. 
     \textit{For example,  adapting a PTM to classify 10 classes of satellite images in EuroSAT~\cite{helber2019eurosat}, requires 1000 training epochs.} This is because, adapting the visual encoder is challenging due to the complexity of high-dimensional visual input and the asymmetric nature of V-L encoders, compared to its text counterpart. 
     One may wonder whether language can overcome this constraint. 
     \item \textbf{Generalizing beyond seen classes.} MR is, by nature, a form of transfer learning. As such, it does not endow an explicit mechanism to generalize beyond what it has seen during adaptation. Recent studies have shown that Vision Language Models (VLMs) have great zero-shot learning capabilities. This would suggest whether one can expect or design an MR algorithm that can benefit from language to generalize beyond its seen classes during adaptation.
     

     \item \textbf{Adaptation without accessing model parameters}
     Our method maintains the original foundation model, thus enabling adaptation via APIs and cases where for ethical constraints, accessing the structure and weights of the foundation model is not possible. Furthermore, preserving the foundation model translates into maintaining its generalization capabilities, a virtue, that algorithms such as MaPLe~\cite{ref_maple} cannot ensure.
\end{itemize}

Our work takes a stride toward addressing the aforementioned questions. In particular, 
we propose \textbf{Language-Grounded Visual Prompting} (LaViP)\footnote{  \href{https://github.com/NilakshanKunananthaseelan/LaViP}{https://github.com/NilakshanKunananthaseelan/LaViP}}
, which enables pixel-space input-aware prompting by leveraging the language integration to adapt downstream tasks (Figure ~\ref{fig:concept}). 
In LaViP, we opt for a low-rank solution to generate language grounded visual prompt.
This substantially reduces the number of parameters to be learned, a quality particularly advantageous in the context of black-box settings. Furthermore, we develop a mechanism to incorporate novel class knowledge without needing to retrain the VPs, enabling our solution to generalize to novel and unseen classes seamlessly. 
To contrast and compare our algorithm against previous art, we have performed a thorough set of experiments, ranging over transfer learning, few-shot learning, and generalization beyond seen classes over 12 recognition datasets. Our empirical study shows that our algorithm consistently outperforms state-of-the-art algorithms by a tangible margin by harnessing the multimodal signals in visual prompts.

To summarize, we have made the following contributions to this work. Firstly, to the best of our knowledge, we are pioneering a  language-grounded MR solution to adapt a visual encoder to downstream tasks. Secondly, we propose a mechanism effectively extending visual prompts beyond seen classes, a feat largely confined to text prompt adaptation. We extensively evaluate and assess our algorithm on three learning paradigms: few-shot learning,  generalization beyond seen classes, and transfer learning.

\section{LaViP}
\label{sec:proposed_method}

Throughout the paper, we denote scalars as $x$, vectors as $\vx$, matrices as $\mX$, and equality by definition as $\triangleq$. 
The Kronecker product between matrix $\mX \in \mathbb{R}^{m \times n}$ and $\mY \in \mathbb{R}^{p \times q}$, denoted by $\mX \otimes \mY \in \mathbb{R}^{mp \times nq}$ is defined as 
\begin{equation}
\resizebox{0.6\columnwidth}{!}{$
\mX\otimes \mY =\\
 \begin{pmatrix}
    x_{11} \mY & \cdots & x_{1n}\mY \\
    \vdots & \ddots & \vdots \\
    x_{m1}\mY & \cdots & x_{mn}\mY 
\end{pmatrix}
$}\;,
\end{equation}
where $a_{ij}$ represents the element in the $i$-th row and $j$-th column of $\mX$. 
Below, we describe \textbf{LaViP}, our input-dependent
visual prompting approach guided by language semantics.
In \textsection~\ref{subsec:lavip1}, we provide a detailed exposition of the underlying rationale of our algorithm and its design. \textsection~\ref{subsec:lavip2} illustrates how LaViP can be transitioned to base-to-novel generalization tasks.

\paragraph{Problem Statement.} 
Given a training dataset $\mathcal{S} = \{(\vx_i, \vy_i)_{i=1}^m\}$ drawn i.i.d. from distribution $\mathcal{D}$, we seek to learn a model to effectively assign input vectors $\vx$ to their corresponding class labels $\vy$, based on the patterns and relationships.
We assume $\vx_i \in \mathbb{R}^{\texttt{H} \times \texttt{W} \times \texttt{C}}$ is an image and $\vy_i \in \Delta^{K-1}$ is its associated label, with $\Delta^{K-1}$ denoting the $K$-simplex. Furthermore, we assume a pre-trained VLM with a visual encoder $\Phi_{\text{vis}}: \mathbb{R}^{\texttt{H} \times \texttt{W} \times \texttt{C}} \to \mathbb{R}^d$
$\Phi_{\text{lan}}: \mathcal{X} \to \mathbb{R}^d$ is at our disposal. Here, $\mathcal{X} \subseteq \mathbb{R}^{d_t}$ denotes the input space of the language encoder, 
in the case of CLIP, a subset of integers defined by its tokenizer.

To achieve this goal, our objective is to generate padding-style visual prompts with a total of ${2\texttt{p}\texttt{C}(\texttt{H} + \texttt{W} - 2\texttt{p})}$ parameters, where $\texttt{C}$ represents channels, $\texttt{H}$ and $\texttt{W}$ denote height and width, and $\texttt{p}$ is the padding size.
Unlike previous visual prompting methods such as VP~\cite{ref_vp}, LaViP adopts an approach where it learns input-specific prompts that are language-grounded.

\begin{figure*}[t]
\centering
\includegraphics[width=0.9\textwidth]{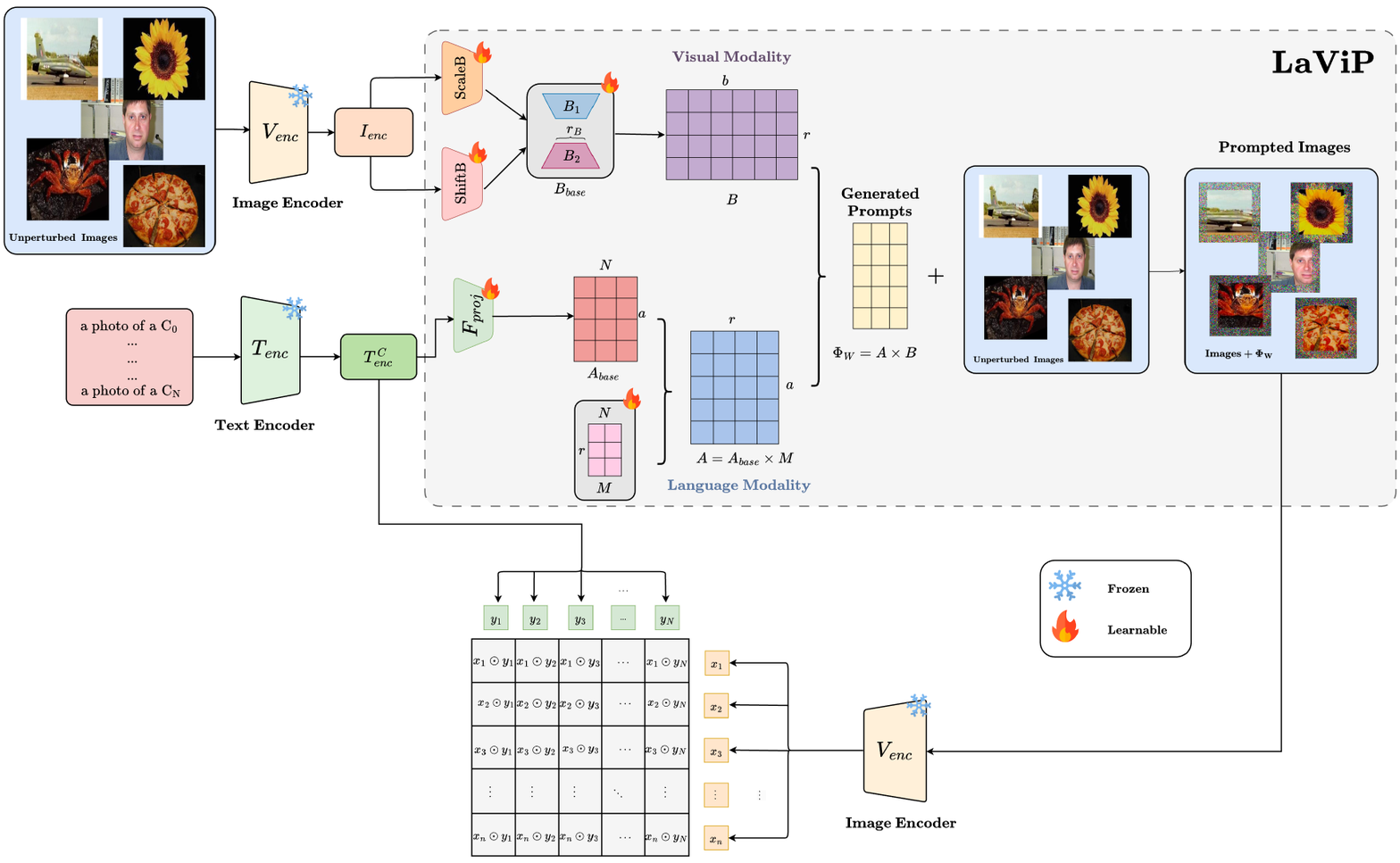} 
\caption{Overview of our proposed Language-Grounded Visual Prompting(LaViP) for VLMs: LaViP utilizes language-grounded input-specific visual programs to reprogram the frozen visual encoder of the CLIP model. LaViP scales and shifts local image encoding and projects global text encoding. The subsequent matrix multiplication of these localized and global projections fosters a mutual synergy between the two modalities, resulting in the generation of adaptive visual prompts.}
\label{fig1}

\end{figure*}
\subsection{Language Grounded Visual Prompts}
\label{subsec:lavip1}
Visual Prompts manipulate the pixel space via learnable parameters and steer the PTMS in any desired direction. While VP made the first contribution to this concept in the context of the pre-trained vision model and VLMs, they overlooked 1) the multimodal nature of VLMs, and 2) the semantic diversity of images. To address these gaps, we propose LaViP, a novel approach that capitalizes on these two important observations. Figure~\ref{fig1} provides an overview of our method. LaViP synergizes complex intricacies in inputs and context expertise, generating language-grounded input-aware visual prompts, which facilitates enhanced modality alignment.

As suggested earlier, the visual prompt for a sample $\vx \in \mathbb{R}^{\mathtt{H} \times \mathtt{W} \times \mathtt{C}}$ is defined as 
$\vnu \in \mathbb{R}^{2\mathtt{C}(\mathtt{H} + \mathtt{W} - 2\mathtt{p})\mathtt{p}}$, which is padded around 
a resized version of $\vx$. We mathematically and with a bit of abuse of notation show this process by: 
\begin{align}
    \tilde{\vx} = \vx \oplus \vnu\;.
    \label{eqn:vp_added_x}
\end{align}

For a VLM such as CLIP, typical values of $\mathtt{H}=\mathtt{W}=224$, and $\mathtt{p} = 28 $, which results in generating ${2\mathtt{C}(\mathtt{H} + \mathtt{W} - 2\mathtt{p})\mathtt{p}}$ parameters for VPs. 

We aim to facilitate the generation of input-specific visual prompts by formulating the process through low-rank matrix decomposition. 
Specifically, we derive two matrices  $\mA \in \mathbb{R}^{a \times r}$,
$\mB\in \mathbb{R}^{r \times b}$ and $\mM\in \mathbb{R}^{K \times r}$. Here, $\mA$ acts as a projection and captures the class semantics of the problem via the language encoder.
Furthermore and as we will show shortly, $\mA$ is obtained from the textual description of all $K$ classes. 
This implies that after training, our algorithm can only store $\mA$ and does not need a language encoder to operate in its nominal form.

On the other hand, the $\mB$ component of the VP is tailored to each image, enabling our method, LaViP, to dynamically adjust its prompts based on the input image it receives. 
This image dependency aligns with the idea that customized guidance can enhance model performance, as previously discussed. We argue that, despite sharing identical class labels, images often exhibit distinct semantic variations. Relying on universal visual prompts limits the model's capacity to adapt effectively to these variations, especially when extending to unseen classes.
The hyper-parameter $r$ controls the rank of $\mA$ and $\mB$, and can be considered as a prior in generating VPs. 
Consequently, we represent the VP as $\vnu = \text{Vec}(\mA \mB)$.
Here, the notation $\text{Vec}(\cdot)$ denotes the process of reshaping a matrix into a vector. 
By adopting this formulation, we reduce the complexity of requiring $\vnu$ from the initially required ${2\mathtt{C}(\mathtt{H} + \mathtt{W} - 2\mathtt{p})\mathtt{p}}$ parameters to merely $r(a + b)$ parameters for each instance. Learning low-rank decomposition of learnable parameters has proven more effective and efficient than finetuning all parameters~\cite {ref_lora}.
Below, we provide a detailed explanation of how $\mA$ and $\mB$ are generated.
\paragraph{Language-Grounded Encoding} Following common practice~\cite{ref_clip,ref_vp}, for all $K$ classes of a downstream, we craft textual descriptions by a template in the form: \enquote{\texttt{a photo of a $\langle class\rangle$}}. 
Then, we obtain the encoding of the VLM language coder for all $K$ prompts as $ \mathrm{T_{enc}} \in \mathbb{R}^{K \times d}$ using:
%
\begin{align}
    \label{eq_txt_enc}
    \mathrm{T_{enc}} = \mathrm{VLM_{TextEncoder}(prompts)}\;.
\end{align}

To enrich the representation, we define $\mA \triangleq \mM \mA_\texttt{base}$. Here, $\mA_\texttt{base}$ is obtained from the semantics $\mathrm{T_{enc}}$ (see Eq.\eqref{eq_txt_enc}) as $\mA_\texttt{base} = f(\mathrm{T_{enc}})$. 
The matrix $\mM$ is learnable, helping the model to gauge the semantics to be incorporated as a part of VPs. 
\paragraph{Image-dependent Encoding} Similar in concept, we formulate the image-dependent part of the VP as a matrix decomposition, 
albeit with some touch-ups. In particular, we propose the following form for constructing $\mB$:
\begin{align}
    \mB \triangleq   \mB_\texttt{scale} \odot \mB_\texttt{base} + \mB_\texttt{shift}\;,
    \label{eqn:b_general_form}
\end{align}
where $\mB_\texttt{scale}, \mB_\texttt{shift} \in \mathbb{R}^{r} $, $\mB_\texttt{base} \in \mathbb{R}^{b \times r}$ and $\odot$ indicates scaling function.
In Eq.\eqref{eqn:b_general_form}, $\mB_\texttt{base}$ is a matrix encoding the visual aspects of the input image and is obtained as:

\begin{equation}\label{eq_mat_mul}
    \mB_\texttt{base} = \mB_\texttt{1} \times \mB_\texttt{2}\;,
\end{equation}
where $\mB_\texttt{1}$  and $\mB_\texttt{2}$  are low-rank decomposition  of $\mB_\texttt{base}$  with rank $r_{\mB}$.
We modulate $\mB_\texttt{base}$ with $\mB_\texttt{scale}$ and $\mB_\texttt{shift}$, which are light-way matrices obtained through simple linear layers. We opt for light-design choices to accelerate image-wise transformation in Eq.\eqref{eqn:b_general_form} without introducing significant computational overhead and provide a convenient way to introduce non-linearity in the process.
Algorithm~\ref{alg:lavip} summarizes the steps involved in our method. 

\begin{algorithm}[tb]
    \caption{LaViP algorithm }
    \label{alg:lavip}
    \textbf{Input}: \textit{Target dataset: } $\mathcal{D}$ with $K$ classes and $X$ images,\\
                    \textit{Pre-trained model}: $F$ ,\\
                    \textit{Prompt learner }: $P$ with trainable parameters. \\
    \textbf{Parameters}: \textit{Parameters of} $P$ :$\mB_\texttt{1},\mB_\texttt{2},\mM,\mathrm{Scale_B,Shift_B}$\\
    \textbf{Output}: \textit{Visual prompt}:$\Phi_{W}$ for the target task.\\
    \begin{algorithmic}[1] 
    \STATE Initialize Parameters:  $\mB_\texttt{1},\mB_\texttt{2}$ and  $\mM$\\
    \STATE Create Visual Projection Matrix: $\mB_\texttt{base} = \mB_\texttt{1} \times \mB_\texttt{2}$\\
    \STATE Construct K textual prompts: $\mathrm{prompts} = \{\texttt{ a photo of a \{i\}}\}_{i=1}^{K}\}$ \\
    \STATE Encode the image and text prompt:\\ $T_{enc},I_{enc} = {F}(x,\mathrm{prompts})$\\
    \STATE Project text encoding : $\mA_\texttt{base} = f(T_{enc})$\\
    \STATE Control the text integration: $\mA = \mM \times \mA_\texttt{base}$\\
    \STATE Scaling and shifting of image encoding: \\$\mB_\texttt{scale} = \mathrm{Scale_B}(I_{enc})$ \\ 
           $\mB_\texttt{shift}= \mathrm{Shift_B}(I_{enc})$ \\
    \STATE Feature-wise modulation: \\$\mB \triangleq   \mB_\texttt{scale} \odot \mB_\texttt{base} + \mB_\texttt{shift}$\\
    \STATE Combine the textual and image-specific information :  $\Phi_{W} = \mA \times \mB$\\
    \end{algorithmic}
\end{algorithm}
    
\subsection{Generalization from Base to Novel Classes}
\label{subsec:lavip2}

In the base-to-novel generalization task, the goal is to evaluate the \textit{generalizability of the model to unseen classes }by training on base classes while evaluating on the base and novel classes \textit{separately}~\cite{ref_cocoop}. 

CoOp~\cite{ref_coop} learns text prompts neglecting input differences, therefore failing to generalize well beyond classes in training data. To alleviate such drawbacks, CoCoOp~\cite{ref_cocoop} proposes image-conditioned text prompts to impute novel class knowledge into prompts, and MaPLe~\cite{ref_maple} injects tokens in both the vision and language branches which efficiently transition the novel class knowledge into prompts. In contrast to these approaches, visual prompt-based techniques lack an efficient means to integrate novel class knowledge. 

We address this limitation by embedding novel-class knowledge into the visual prompts on the fly and without the need for retraining.
The Kronecker product encapsulates information, eliminating the necessity for additional learning~\cite{ref_kronecker1,ref_kronecker2,ref_kronecker3,ref_kronecker4}.
The underlying idea is to employ the similarity between novel classes and base classes to refine $\mA$. 
Recall that $\mA = \mM \mA_\texttt{base}$. This can be understood as $\mA$ being a linear combination of the semantic information captured in 
$\mA_\texttt{base}$. In the presence of novel classes, we first encode a notion of similarity between base and novel classes by 

\begin{align}
\resizebox{0.8\columnwidth}{!}{$
\mathrm{T_{enc}^{K_{novel}}} \otimes \mA =
 \begin{pmatrix}
    t_{11} \mA & \cdots & t_{1d}\mA \\
    \vdots & \ddots & \vdots \\
    t_{K_{novel}1}\mA & \cdots & t_{K_{novel}d}\mA
\end{pmatrix}
$}
\;,
\end{align}
where $\otimes$ denotes the Kronecker product.

The resulting product will exist in $\mathbb{R}^{(aK_{novel}) \times (rd})$, representing the projection of each novel class on all the base classes. To obtain a compact and coherent embedding representation between base and novel classes,  we transform this class-wise projection into $\mathbb{R}^{a \times K_{novel} \times r\times d}$. Subsequently, we compute the mean along VLM features($d$) and the number of novel classes ($K_{novel}$). The averaging operation serves to align the base classes with a more unified representation of novel classes.


\section{Related Works}

In this section, first, we will introduce VLMs and their constraints for adapting to new tasks, then we will discuss existing prompt learning methods, and finally, we will explore how MR is used in repurposing PTMs for diverse domains tasks.
\begin{table*}[t]
\small
\centering
\resizebox{0.85\textwidth}{!}{
\begin{tabular}{l|cccccccccccc|c|c}
    \toprule
    
    Method & Caltech & Pets & Cars & Flowers & Food & Aircraft & SUN & DTD & EuroSAT & RESISC & CLEVR & UCF &\textit{Avg.} & \textit{Win}\\
    \hline
    \\
    ZS\cite{ref_clip} & 89.3 & 88.9 &65.6 & 70.4 & \textbf{89.2} & 27.1 & 65.2 & 46.0 & 54.1 & 65.5 & 23.4 & 69.8 &62.75&1\\
    
    VP\cite{ref_vp} & 94.2 & 90.2 & 66.9 & 86.9 & 81.8 & 31.8 & 67.1 & 61.9 & \textbf{90.8} & 81.4 & 40.8 & 74.2 &71.26&1\\
    \midrule
    \textbf{LaViP(Ours)} & \textbf{95.0} & \textbf{91.2} & \textbf{77.8}  &  \textbf{96.3} & 82.5 & \textbf{43.2} & \textbf{71.1}& \textbf{68.8} & 86.1 &\textbf{85.6} & \textbf{46.5} & \textbf{81.3}&\textbf{74.59}&10\\
    \bottomrule

\end{tabular}
}
\caption{Comparison with visual prompting method on few-shot transfer learning. LaViP learns language-driven input-aware visual prompts and exhibits robust performance on 10 in 12 recognition datasets with training accelerated by \textit{more than  $3 \times$.} \textit{Win} indicates how many cases LaViP outperforms previous methods.}
\label{tbl:fsl}
\end{table*}

\subsection{Pretrained Vision-Language models:}

VLMs such as CLIP~\cite{ref_clip}, ALIGN~\cite{ref_align}, Flamingo\cite{ref_flamingo}, Flava~\cite{ref_flava} and LiT~\cite{ref_lit} have demonstrated exceptional performance on a variety of tasks, including few-shot and zero-shot image recognition. These models learn to align the vision-language representations on a web-scale training dataset.
Although pre-trained models offer a strong foundation for a wide range of tasks, efficiently adapting them to downstream tasks is still a challenging research problem. 
The difficulty is exacerbated when the downstream task requires specialized context, interpretable representations, or access to the model is forbidden~\cite{ref_clipcap,ref_ft_prob1,ref_ft_prob2,ref_aar_prompting}.
Furthermore, Kumar \etal showed finetuning overparameterized models can yield detrimental results compared to linear probing (i.e., tuning the head while keeping lower layers frozen) when addressing out-of-distribution downstream tasks~\cite{kumar2021fine}. 


\subsection{Prompt Learning in VLMs}

Standard finetuning and linear probing are common approaches to adapting VLMs to downstream tasks. However, such finetuning causes detrimental results due to the loss of embedded knowledge and poor adaptation techniques~\cite{ref_finetuning2}.
There is a significant body of work in natural language processing (NLP) that focuses on learning effective prompts to adapt a large language model to downstream tasks~\cite{ref_nlp1,ref_nlp2,ref_nlp3,ref_nlp4}.
Inspired by the success of prompt learning in NLP, several recent studies explored prompt learning methods in the context of large-scale VLMs.
Visual Prompt Tuning(VPT) learns the prefix prompts in encoder layers or embedding layer~\cite{ref_vpt},
while ~\cite{ref_maple} proposes injection of learnable tokens in both vision and text encoder layers and couples them with a learnable function.
Visual Prompting(VP) investigated input pixel space prompt tuning for pre-trained vision and VLMs~\cite{ref_vp}. VP learns a fixed input agnostic perturbation and attaches it to the original images, hence adapting a pre-trained model to new tasks without modifying the model parameters.~\citeauthor{ref_vp_inpaint} uses the inpainting method as visual prompting. Context Optimization (CoOp)~\cite{ref_coop} optimizes a set of context vectors for the text encoder of CLIP, while Conditional Context Optimization (CoCoOp)~\cite{ref_cocoop} generalizes CoOp to unseen classes by conditioning the text prompt optimization on image instances. ~\cite{ref_multimodality} suggests cross-modal adaptation by repurposing class names as one-shot training examples, ~\cite{ref_prompt_dist} proposes an ensemble of learnable prompts.
~\cite{ref_llm_vis,ref_cafo,ref_using} showcase how language can be effectively employed to strengthen the adaptation of pre-trained vision models to novel domains.

We argue that generating input-agnostic prompts with unimodal knowledge is a suboptimal approach. Considering the large-scale pre-training of VLMs, prompting methods should adeptly utilize the embedded multimodal knowledge to efficiently address new tasks.  Further, we underscore the importance of prompting methods being agnostic to the underlying architecture of PTMs. For instance,  VPT and MaPLe have successfully adapted ViT encoders through prefix learning. However, these methods lack comprehensive evidence of how their solutions perform across diverse backbone architectures.

\begin{table*}[t]
\footnotesize
    \centering
    \renewcommand{\arraystretch}{1.1}
\renewcommand{\tabcolsep}{2.0mm}
    \resizebox{0.85\textwidth}{!}
    {
    \begin{tabular}{l|ccc|ccc|ccc|ccc|ccc}
        \toprule
        \multicolumn{1}{c|}{} & \multicolumn{3}{c|}{CLIP} & \multicolumn{3}{c|}{CoOp} & \multicolumn{3}{c|}{CoCoOp} & \multicolumn{3}{c|}{MaPLe}&\multicolumn{3}{c}{\textbf{LaViP (Ours)}} \\
        
         \cmidrule(r){2-4} \cmidrule(lr){5-7} \cmidrule(lr){8-10} \cmidrule(lr){11-13}\cmidrule(l){14-16}
        Dataset & \textbf{Base} & \multicolumn{1}{c|}{\textbf{Novel}} & \multicolumn{1}{c|}{\textbf{HM}} & \textbf{Base} & \multicolumn{1}{c|}{\textbf{Novel}} & \textbf{HM} & \textbf{Base} &\multicolumn{1}{c|}{\textbf{Novel}} & \textbf{HM} & \textbf{Base} & \multicolumn{1}{c|}{\textbf{Novel}} & \textbf{HM}& \textbf{Base} & \multicolumn{1}{c|}{\textbf{Novel}} & \textbf{HM} \\

        \midrule
        Caltech101 &96.84 &94.00 & 95.40 &\textbf{98.00}&89.81&93.73&97.96 &93.81& 95.84&97.74&\textbf{94.36}&\textbf{96.02}& 97.63 &93.45 &95.49\\
        
        DTD & 53.24& \textbf{59.90}& 56.37&79.44 &41.18& 54.24 & 77.01& 56.00& 64.85& \textbf{80.36} &59.18& \textbf{68.16}& 80.05& 58.01&67.27\\ 
        
        EuroSAT& 56.48 &64.05& 60.03 &92.19& 54.74& 68.69&87.49& 60.04 &71.21&\textbf{94.07}& 73.23&82.35 &92.53& \textbf{82.31} &\textbf{87.12} \\ 
        
        FGVCAircraft&27.19& \textbf{36.29} &31.09 &\textbf{40.44} &22.30 &28.75 &33.41 &23.71& 27.74 &37.44& 35.61 &\textbf{36.50}& 37.25&34.03 &{35.57}\\
        
        Food101& 90.10 &91.22 &90.66&88.33& 82.26& 85.19 & {90.70} &{91.29}& {90.99} &\textbf{90.71} &\textbf{92.05 }&\textbf{91.38}& 86.19 &91.28 &88.66 \\ 
        
        OxfordPets&91.17&97.26& 94.12&93.67 &95.29& 94.47& {95.20} &{97.69}&{ 96.43} &\textbf{95.43}& \textbf{97.76}& \textbf{96.58}& 92.45&97.22 &94.78 \\ 
        
        SUN397& 69.36 &75.35& 72.23 &{80.60} &65.89& 72.51 & 79.74 &{76.86}& {78.27 }&\textbf{ 80.82}& \textbf{78.70}& \textbf{79.75}& 76.47 &73.25&74.82 \\ 
        
        Flowers102& 72.08& \textbf{77.80} &74.83&\textbf{97.60} &59.67& 74.06  & 94.87 &71.75& 81.71  &95.92 &72.46& 82.56  & {96.96} &{76.34} &\textbf{85.25}\\
        
        UCF101& 70.53 &{77.50} &73.85& \textbf{84.69} &56.05 &67.46& 82.33 &73.45 &77.64  & 83.00& \textbf{78.66}& \textbf{80.77} & {83.83} &76.46 & {79.97} \\ 
        
        StanfordCars&63.37 &\textbf{74.89}& 68.65  & \textbf{78.12} &60.40 &68.13  &70.49&73.59&72.01 &72.94 &74.00& 73.47 & {73.63}&{74.63}&\textbf{74.13} \\ 

        \midrule
        \textit{Average}&69.04&74.83&71.72&\textbf{83.31}&62.76&70.72&80.92&71.82&75.67&{82.84}&75.60&\textbf{78.75}&81.7&\textbf{75.70}&78.31\\
        \bottomrule
    \end{tabular}
    }
    \caption{Performance of LaViP  on base-to-novel generalization across 10 recognition datasets. LaViP demonstrates competitive generalization performance over CoOp and CoCoOp with an absolute gain of \textit{2.64\%}.HM indicates the harmonic mean of base class accuracy and novel class accuracy.}

\label{tbl:b2n}
\end{table*}

\subsection{Model reprogramming}
By deriving motivation from adversarial attacks ~\citet{ref_ar} proposed Adversarial Reprogramming(AR) to repurpose a pre-trained model to perform on a new domain. This led to a new learning paradigm called model reprogramming (MR) for transfer learning.
 
We provide some notable examples below. ~\citet{ref_mr1} repurposed a language model to predict biochemical sequences;  ~\citet{ref_bar} proposed BAR to reprogram an ImageNet model for complex bio-medical under a black-box setting; ~\citet{ref_mr2}  used an attention-based RNN speech model for low-resource spoken command recognition;
~\citet{ref_mr3} reprogrammed a speech model for time-series prediction and ~\citet{ref_mr4} reprogrammed a vision model to classify text sentences and DNA sequences. 
~\cite{ref_blackvip} extended BAR by generating input-aware visual prompts through an external encoder-decoder model for limited data recognition. 

To the best of our knowledge, we are pioneering to design of language-grounded visual prompts to reprogram the visual encoder of a VLM. In contrast to previous MR methods that primarily focused on repurposing PTMs using unimodality, our contribution lies in harnessing the power of multimodality to enhance context knowledge during adaptation.


\section{Results}
We first provide the experimental setup in \textsection~\ref{experimental}. Next, \textsection~\ref{fsl} presents the comparison between LaViP and previous methods. \textsection~\ref{b2n} provides the result for the base-to-novel generalization task and \textsection~\ref{fullset} provides the results for whole-dataset training.

\subsection{Experimental Setup}
\label{experimental}
We extensively evaluate LaViP capability on 12 benchmark datasets (refer Appendix~\ref{supp:datasets}) under three distinct scenarios.\\
First, its transferability in limited data settings is assessed through few-shot learning, where it learns from 16-shots for training and 4-shots for validation. Next, its generalizability is examined by evaluating its ability to learn from base classes and apply that knowledge to unseen novel classes. Finally, we use the full dataset for training, testing and validation. In this paper, we use CLIP ViT-B/16 ~\cite{ref_clip} for few-shot learning and base-to-novel generalization, and CLIP ViT-B/32 for whole-set training as the pre-trained VL model due to its strong zero-shot generalization capability. More details are provided in Appendix~\ref{supp:experiments}.

\subsection{Few-shot learning}
\label{fsl}
Table~\ref{tbl:fsl} presents the performance of LaViP in a few-shot transfer setting across 12 recognition datasets. We compare our results against CLIP Zero-shot (ZS), and the previous pixel-space reprogramming method.
 LaViP outperforms ZS on 11 datasets, exhibiting a substantial gain of 11.84\%. Additionally, when comparing to VP\cite{ref_vp} on 11 datasets,  achieving a gain of 3.3\% in performance and more than a $3\times$ faster convergence.
Furthermore, Table~\ref{tbl:fsl} shows that when the domain shifts from generic to rare concepts, LaViP consistently demonstrates higher performance in comparison to CLIP. This highlights the effectiveness of incorporating language guidance in enhancing modality alignment, particularly in cases where concepts can be explicitly described.

\subsection{Base-to-Novel Generalization}
\label{b2n}
Table~\ref{tbl:b2n} presents the performance of LaViP in the base-to-novel generalization setting, evaluated across 10 recognition datasets. We compare LaViP against a lineup of benchmarks, including CLIP Zero-shot(ZS), CoOP, CoCoOp and MaPLe.
Relative to CoCoOp, LaViP exhibits a stronger performance across both base and novel concepts, yielding absolute gains of 0.0.78\% and 3.88\% respectively. With the context-aware knowledge diffused through Kronecker product, LaViP as a strong competitor surpasses CoCoOp in 6/10 datasets and slightly trails in two datasets. When taking into account both base and novel classes, LaViP shows an absolute average gain of with gain of 2.64\% compared to CoOp and CoCoOp.

MaPLe, the current SOTA has outperformed in many studied datasets. However, unlike MaPLe, other algorithms maintain the original foundation model, thus enabling adaptation via APIs and cases where for ethical constraints, accessing the structure and weights of the foundation model is not possible. Furthermore, preserving the foundation model translates into maintaining its generalization capabilities, a virtue, that algorithms such as MaPLe cannot ensure.
LaViP trailed MaPLe by only 0.44\% in performance, showcasing a competitive performance and even marginally outperforming on classifying novel classes.  

In comparison with CLIP on novel classes, CoCoOp improves 3/10 classes, leading to a decrease in the average novel accuracy from 74.83\% to 71.82\%. LaViP only improves accuracy in 2 out of 10 datasets compared to CLIP for new classes. However, it positively impacts the average accuracy, elevating it from 74.83\% to 75.70\%. This sustains its position as a robust competitor. CoOp exhibits limited generalization capability to novel classes, a deficiency that CoCoOp strives to address by contextualizing text prompts based on image instances, and shows substantial improvement in novel class recognition. However, it manages to outperform 2/10 base classes with a decrease in average performance of 2.39\%. LaViP's language integration exhibits competitive performance in the base class, with only a 1.4\% decrease in average performance. 
Despite marginal improvements compared to CoCoOp, it's vital to recognize the dissimilarity between high-dimensional, variable-rich image data and structured text. This divergence affects learning speed and effectiveness, especially with limited data. 
Given the complexities inherent in the visual domain, an approach must be parameter-efficient and context-aware. This dual requirement aligns with the fundamental characteristic of LaViP.

Moreover, from  Table~\ref{tbl:b2n} we can conclude that as the domain shift increases from the pretraining dataset, LaViP exhibits increasing performance compared to CLIP, CoOp and CoCoOp. This emphasizes the impact of language context in designing visual prompts.

\begin{table}[h]

\centering

\resizebox{\columnwidth}{!}{

\begin{tabular}{l|ccccccccc|c}
    \toprule
    Method & Pets & Flowers & Food & SUN & DTD & EuroSAT & RESISC & CLEVR & UCF  &\textit{Avg.} \\
    \midrule
    ZS &88.3 &67.4 & 85.2 &62.6 & 44.4& 42.2 & 56.6 & 25.8 & 65.2& 59.75\\
    
    CLIP + LP&89.2 & 96.9& \textbf{84.6} &\textbf{75.0} & \textbf{74.6} &  95.3 & \textbf{92.3} & 66.0 & \textbf{83.3}&84.13\\
    VP &85.0 & 70.3 & 78.9&60.6 & 57.1 & \textbf{96.4} & 84.5 & \textbf{81.4} & 66.1&75.72\\
    \midrule
    \textbf{LaViP(Ours)} &\textbf{89.6}&96.7 &83.2 &71.5 &72.9&96.3&91.0&67.0&81.9&80.99 \\
    \bottomrule
\end{tabular}
}
\caption{Performance across 9 recognition dataset using CLIP Zero-shot(ZS), Linear Probe(LP), Visual Prompting(VP) and LaViP with ViT-B/32 backbone.}
\label{tbl:full}
\end{table}
\subsection{Full-dataset Learning}
\label{fullset}
The summarized findings of full-dataset training are presented in Table~\ref{tbl:full}.
To provide a comprehensive evaluation, we draw comparisons between our results and those of CLIP Zero-shot (ZS), CLIP Linear Probe (LP), and VP, all using the ViT-B/32 CLIP model with the same hyperparameters as those used in few-shot learning.  It indicates that LaViP consistently outperforms VP by a notably wider margin across 7/9 recognition datasets. This substantial improvement is coupled with an optimization process that is three times more efficient.
In contrast with the results obtained from CLIP Linear Probe, VP shows enhancement in 2 out of 9 datasets, albeit accompanied by an average accuracy decrease of 8.91\%. Comparatively, while LaViP only enhances performance in 1 out of 9 datasets, it achieves an average accuracy drop of 3.14\%. These observations highlight the robust improvements brought about by improved visual prompting of LaViP.


\section{Ablation Studies}
\subsection{Learning in gradient-free environment}
We adapt our algorithm in a gradient-free environment to understand the effectiveness of language integration. 
We proceed to evaluate \textbf{BlackLaViP}, the gradient-free variant of LaViP, by employing the SPSA algorithm ~\cite{ref_spsa1,ref_spsa2}. Table~\ref{tbl:black_lavip} presents the performance of BlackLaViP on few-shot learning on 10 recognition datasets.
BlackVIP~\cite{ref_blackvip} uses SPSA with an external model to generate input-aware prompts and exhibits effectiveness on 4 out of 10 datasets. Though BlackLaViP doesn't outperform BlackVIP, an intriguing observation emerges from our experiment. Remarkably, BlackLaViP attains 95\% of the performance of BlackVIP with a convergence rate that is over \textit{$15 \times$} faster. Implementation details are provided in Appendix~\ref{supp:black-box-learning}. 

\subsection{Impact of hyperparameters $(a, b, r)$}

VP requires generating a padding of size $\theta={2\texttt{p}\texttt{C}(\texttt{H} + \texttt{W} - 2\texttt{p})}$ as the visual prompt. We propose a low-rank formulation to generate the prompt, which efficiently reduces the generator size by a factor of 4 when $r=32$, 2 when $r=64$, and 1.2 when $r=96$. The parameter $r$ (rank of matrices in generating the prompt) can be understood as an inductive bias, regularizing the design. Empirically, we observed that   LaViP performed robustly if $r$ was chosen within a reasonable range (not too small, \eg $r \in [16,96]$). Furthermore, LaVIP robustly performs for varying $(a, b)$ which creates padding of size $p=20$ to $p=50$. Additional results are provided in Appendinx{~\ref{supp:ablation}}.

\begin{table}[t]
\centering
\resizebox{\columnwidth}{!}{
\begin{tabular}{l|cccccccccc|c}
    \toprule
    Method & Caltech & Pets & Cars & Flowers & Food & Aircraft & DTD & RESISC & UCF &SUN&\textit{Avg.}\\
    \midrule
    
    ZS(CLIP) & 89.3 & 88.9 &\textbf{65.6} & 70.4 & \textbf{89.2 }& \textbf{27.1 }& 46.0 & \textbf{65.5} &  \textbf{69.8}  &62.6&67.44\\
    VP w/SPSA-GC&89.4&87.1&56.6&67.0&80.4&23.8&44.5&61.3&64.6&61.2&63.59\\
    
     BAR & \textbf{93.8} & 88.6 & 63.0 & \textbf{71.2} & 84.5 & 24.5 & \textbf{47.0} &  65.3 &  64.2 &62.4&66.45\\

    BlackVIP & 93.7 &\textbf{ 89.7} & \textbf{65.6} & 70.6 & 86.6 & 25.0 & 45.2 & 64.5 & 69.1&\textbf{64.7}&67.47 \\
    \midrule
    \rowcolor[gray]{0.9}
    \textbf{BlackLaViP (Ours)}& 92.3&89.3&63.3&68.8&84.6&23.9&46.1&61.8&65.6 &62.2&65.77\\
    
    \%&98.5&99.6&96.5&97.5&97.7&95.6&101.2&95.8&95.1&96.1&97.49\\
    \bottomrule

\end{tabular}
}
\caption{Comparison with state-of-the-art methods on few-shot transfer learning in a black-box setting. \% indicates percentage of BlackVIP score achieved with\textit{ more than $15\times$} faster optimization.}
\label{tbl:black_lavip}
\end{table}

\section{Discussion}

LaViP, a novel approach to visual prompting, harnesses the power of language to enhance pre-trained models without the need for invasive finetuning. By merging textual knowledge into input prompts, LaViP steers models towards desired tasks, surpassing the limitations of previous methods in both accuracy and optimization. The few-shot capability is a direct result of preserving the foundation model. By aligning image and text via visual prompting and without altering the latent space of the foundation model, we capitalize on the generalization capability of the model. 

Its versatility shines across diverse tasks, requiring no individual finetuning efforts, and its privacy-preserving nature makes it ideal for interacting with APIs and proprietary software. The reprogramming methodology studied in this can work to provide increased user control over bias and fairness issues in pretraining. However, low-resolution images and highly diverse datasets present challenges. We hypothesize that the observed characteristic is due to context tokens failing to grasp semantic content or capture the full spectrum of classes. LaViP's performance is inherently influenced by the context tokens present in the prompt template. This naturally gives rise to the question: \textit{What advantages does learning text prompts offer in comparison to employing manually crafted templates in LaViP?}. Future research could delve into the direction of learning multimodal prompts with mutual synergy.

\section{Conclusion}

Adaptating large-scale VLMs(\eg CLIP~\cite{ref_clip}) for new tasks is a challenging research problem due to a large number of tunable parameters. Despite stemming from distinct motivation, prompt learning and model reprogramming provide an efficient and scalable approach to drive VLMs to downstream tasks. To this end, existing visual prompting approaches learn input-agnostic prompts through unimodal knowledge. The perceptual diversity of the image domain makes a difficult to repurpose visual encoders in VLMs compared to text encoders, and these approaches require an external world model to provide context or a large number of iterations. Our work counters these assumptions by leveraging embedded multimodal knowledge within VLMs. Our approach seamlessly integrates these multimodal representations to generate adaptable visual prompts, thereby enhancing performance without compromising. Further, we propose an efficient strategy for generalizing visual prompting methods to unseen classes.
Our method improves the few-shot transfer learning, generalization towards novel concepts and full-set transfer learning with varying domain shifts compared to the pretraining dataset.

\bigskip
\fontsize{9.0pt}{10.0pt} \selectfont
\bibliography{main}

\clearpage
\appendix
\fontsize{10.0pt}{11.0pt} \selectfont

\section*{Supplementary Material}

This section contains supplementary material that provides additional details for the main paper and further experimental analysis. This section outlines the contents in the following order.
\begin{itemize}
    \item Preliminary studies (Appendix~\ref{supp:preliminary}).
    \item Experimental setting (Appendinx~\ref{supp:experiments})
    \item Optimizing in a gradient-free environment (Appendix~\ref{supp:black-box-learning}).
    \item Additional ablation experiments (Appendix~\ref{supp:ablation}).  
\end{itemize}


\section{Preliminary}
\label{supp:preliminary}
We first provide a brief introduction to why visual prompting is challenging in VLMs compared to text prompting, then we discuss the mechanism of CLIP and finally discuss how visual prompting works.
 \subsection{Why Visual prompting is challenging in CLIP?}

\begin{figure}[h!]
\centering
\includegraphics[width=\columnwidth]{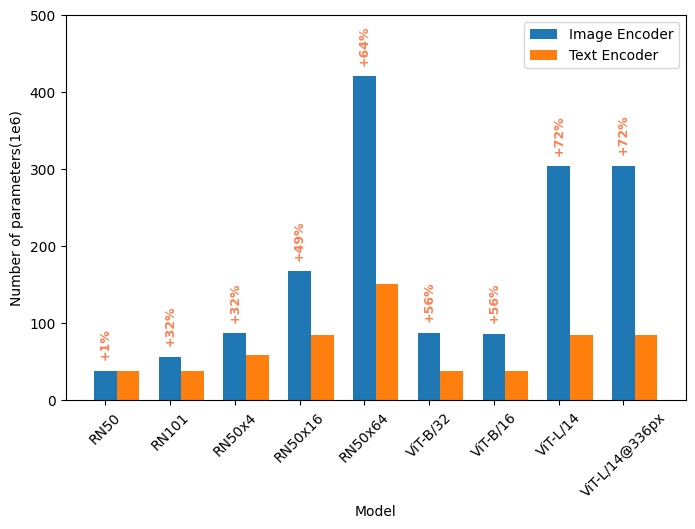} 
\caption{Number of parameters in visual and text encoders of CLIP variants~\cite{ref_clip}}
\label{fig2}
\end{figure}

Images are often high-dimensional data with rich visual cues and more variability, whereas text data can be more structured and symbolic. This nature of data can affect how fast and effectively models learn and optimize, particularly if the training data is limited. By the asymmetric design of CLIP images and text encoders, their parameters influence the capacity of adapters to learn from a few data. With a large number of parameters of image encoders in CLIP as shown in Figure~\ref{fig2}, they may require more data and time to learn complex patterns and optimize them effectively. Once pre-trained, finetuning the vision encoder fully or partially has the potential risk of failing to preserve the learned patterns.

\begin{table}[t]

\centering
\resizebox{0.6\columnwidth}{!}{
\begin{tabular}{l c c c}
    \toprule
   Method &Prompt& Params &  \%CLIP \\
   \midrule
   CoOP& L&2048 & 0.002\\
   CoCoOP& L&35360&0.03\\
   VP&V& 69840&0.06\\
   VPT&V&15.51 M & 12.51\\
   MaPLe&V-L& 3.5 M & 2.83\\
   \midrule
   \textbf{LaViP} \textit{(Avg)}& V-L &0.221 M & 0.18\\
   \hline
   
\end{tabular}
}
\caption{Number of trainable parameters used in existing methods} 
\label{table:params}
\end{table}

Table~\ref{table:params} shows several learnable parameters used by different prompt learning methods. We note that, despite tuning a few parameters, language prompt learning methods such as  CoOP and CoOP exhibit strong performance on few-shot learning compared to visual prompting methods(VP and VPT). Therefore, to use the visual prompting method a new design paradigm that can promote the alignment between visio-language modalities is essential. Thereby, we propose visual prompting by leveraging the language knowledge embedded in CLIP.

\subsection{CLIP}

Our approach is based on CLIP which is a pre-trained vision language model that uses vision and text encoder to align the two input modalities.  We evaluate LaViP using ViT-based CLIP to make a fair comparison with previous visual prompting methods. LaViP optimizes the input transformation similar to MR, therefore it can be used with vision-only models provided text embeddings and output are available. Below, we expand our discussion using ViT-based CLIP architecture.
\subsubsection{Image Encoding:}
The image encoder $V_{enc}$ consists of K transformer layers $\{V_{enc^{i}}\}_{i=1}^{N}$ input dimension of $d_v$ and output dimension of $d_{vl}$. The input image is split into M fixed-size patches using a CNN layer, and then the patches are projected to linear embeddings $e_0 \in \mathrm{R}_{M \times d_{v}}$. 
The embeddings from the previous layer are passed on to the next transformer layer, along with a learnable class token $c_i$, and then processed sequentially through N transformer layers. Image patches are treated the same way as tokens described in [Transformer.] \\For $i=1,2,\cdots,K$
\begin{equation}\label{eq_Venc}
    [c_i,E_i] = V_{enc^{i}}([c_{i-1},E_{i-1}])
\end{equation}
The class token of the final transformer layer,$c_N$ is projected to a shared vision-language embedding space via a $\mathrm{ImageProj}$ to form $I_{proj}$.

\begin{equation}\label{eq_img_proj}
    I_{proj} = \mathrm{ImageProj}(c_k) ;\;\;\;I_{proj} \in \mathrm{R}_{d_{vl}}
\end{equation}
\subsubsection{Text Encoding:}

CLIP text encoder $T_{enc}$ consist of K transformer layers with input dimension of $d_l$ and output dimension of $d_{vl}$. It first tokenizes the words to n tokens and then maps them to corresponding token embeddings  $W_{0} \in \mathrm{R}^{N \times d_{l}}$. 
Token embeddings from the previous layer are passed to the next transformer layer and sequentially processed through K transformer layers.
\\For $i=1,2,\cdots,K$
\begin{equation}\label{eq_Tenc}
    [W_i] = T_{enc^{i}}([c_{i-2},E_{i-1}])
\end{equation}

The final token of the $T_{enc^{K}}$ transformer layer is projected into shared vision-language embedding space via a $\mathrm{TextProj}$ to form $T_{proj}$.

\begin{equation}\label{eq_txt_proj}
    T_{proj} = \mathrm{TextProj}(c_k)  ;\;\;\;T_{proj} \in \mathrm{R}_{d_{vl}}
\end{equation}
\subsubsection{Aligining Vision and Language projection:}

Once the image projection $I_{proj}$ and language projection $T_{proj}$ are obtained, they are $\ell_{2}$-normalized and a similarity score$(sim(\cdot))$ is computed between the normalized projections. Both the contrastive learning processes (i.e. $V\rightarrow L$ and $L\rightarrow V$ ) are regarded as symmetrical operations and they are used jointly optimized during the training phase.

\subsubsection{Zero-shot Classsification}

For the zero-shot classification task, 
A set of hand-crafted prompts are used:\enquote{\texttt{a photo of a $\langle class\rangle$}})  with class labels $y \in \{1,2,\ldots C\}$.
To predict the output, the $sim(\cdot)$ score between the image instance and the prompts from all the class labels is calculated. and the final score is adjusted using a scaling temperature parameter $\tau$. The predicted output $\hat{y}$ is the one with the highest $sim(\cdot)$ score.

\begin{equation}\label{eq_pred}
    p(\hat{y} | I_{proj}) = \frac{\exp(sim(I_{proj},T_{proj}^{\hat{y}})/\tau)}{\sum_{i=1}^{C} \exp(sim(I_{proj},T_{proj}))}
\end{equation}

\subsection{Revisiting Visual Prompting }

Studies in NLP explored prompt learning as a method to calibre the inputs of pre-trained language models to new tasks. This motivates the use of prompt learning methods to exploit the knowledge of VLMs. Studies such as CoOP learn to adapt to the language branch, while VP learns to adapt to the vision branch.\newline
Below, we discuss the shared rationale behind these concepts in adapting pre-trained models.
Given a frozen PTM $F$ and a target dataset $\mathcal{D} = \{(x_1,y_1),\ldots,(x_k,y_k)\}$, the objective is to train a prompt learner ($P$) such that it generates perturbation in input space $\Phi_{W}$, parameterized by learnable parameters $W$ of $P$.
The generated perturbation is attached to the input before feeding to the pre-trained model.

\begin{equation}\label{eq_p_add}
    X_{prompted} = X + \Phi_{W} 
\end{equation}

During training, the P is learned to maximize the likelihood of the correct label y,
\begin{equation}
    \max\limits_{W} \Phi_{W}(y|X+\Phi_{W})
\end{equation}
while the gradient updates are applied only to $P$. During the evaluation, the optimized parameters $W^{*}$ of P are used to generate a prompt to the input and added to all the input instances.
\begin{equation}\label{eq_p_test}
    X_{test} = \{x_1 + \phi_w,\ldots,x_m + \phi_w\}
\end{equation}

Given the perceptual variability of images and semantic abstraction of textual descriptions, we assert that prompt learning methods need to be adaptable for input disparity.


\section{Experimental Setting}
\label{supp:experiments}


\subsection{Backbone Model}
\label{supp:backbone}
The primary focus of this work is directed towards achieving robust adaptation of pre-trained V-L models to diverse downstream tasks. To conduct our experiments, we utilized the publicly available vision language model CLIP~\cite{ref_clip}. To make a fair comparison with previous arts for the underlying image encoder, we chose the ViT-b/16~\cite{ref_vit} with dimensions $d$ = 512, as our default option. Throughout the training phase, we left the entire parameters of CLIP architecture and refrained from making any structural adjustments. Our focus is solely on optimizing the prompt learner $P$ externally.

\subsection{Baseline Methods}
\label{supp:baseline}
1) \textit{CLIP Zero-Shot(ZS)}: CLIP learns a robust shared latent space by aligning vision-language modalities through contrastive learning on nearly 400 million image-text pairs, enabling an excellent generalization ability. We aim to improve CLIP's zero-shot performance for new tasks. 2) \textit{VP}: \cite{ref_vp} proposed VP to adapt VLMs by learning pixel space prompts. It uses universal visual prompts for all images.
Our algorithm expands the VP in two significant ways: 1) Creating prompts that are tailored to the input and 2) Using language supervision to enhance the synergy between the vision-language modalities.

\subsection{Datasets}
\label{supp:datasets}
We follow the CoOP~\cite{ref_coop}, VP~\cite{ref_vp} and BlackVIP~\cite{ref_blackvip} and evaluate our method on different image recognition datasets across a diverse range of tasks. This includes:
General images Caltech101~\cite{rref_caltech};  fine-grained datasets, OxfordPets~\cite{ref_pets}, StanfordCars~\cite{ref_cars}, Flowers102~\cite{ref_flowers}, Food101~\cite{ref_food}, and FGVCAircraft~\cite{ref_fgvc}; a scene recognition dataset SUN397~\cite{ref_sun}; an action recognition dataset UCF101~\cite{ref_ucf101}; a texture dataset DTD~\cite{ref_dtd} and a satellite image datasets EuroSAT~\cite{ref_eurosat}; an remote sensing scene recognition dataset RESISC45~\cite{ref_resisc}, visual reasoning dataset CLEVR~\cite{ref_clevr}, and low-resolution datasets  SVHN~\cite{ref_svhn},CIFAR10 and CIFAR100~\cite{ref_cifar}.
We conduct our experiments using data split provided by~\cite{ref_coop,ref_blackvip}.
For few-shot learning experiments, we use 16-shot for the train set, 4-shot for the validation set, and the whole test set. 

\subsection{Implementation Details}
\label{supp:implementation}

We use the default padding of size 28 for all our experiments based on the assessment of various padding sizes and CoOP style transformation to resize the image for CLIP.
All the experiments run consist of 300 epochs run on a single NVIDIA RTX A5500 GPU, with batch size 64 for few-shot learning and 128 for full-set transfer learning.
For the sake of simplicity in our few-shot transfer learning experiments, we set $r=K$ and $r_{\mB}=32$ for $P$ parameters when dealing with datasets featuring over 15 classes. For other datasets,  we set $r=64$ (including SUN397). In the context of base-to-domain generalization experiments, both $r$ and $r_{\mB}$ are set at 32. 
We used SGD with a learning rate of 1.0 for all the datasets, except for StanfordCars, where we utilize a rate of 0.1. The final checkpoint is consistently used for evaluating our algorithm's performance on the test set throughout all experiments. To ensure reproducibility in our few-shot training experiments, performance evaluation relies on calculating average accuracy across three distinct random seeds. 


\section{Optimizing in a Gradient Free Environment}
\label{supp:black-box-learning}

It is important to acknowledge that accessing the parameters of PTMs is not always attainable, which creates a challenge for adapting them to specific downstream tasks. Adapting in a gradient-free environment is challenging because efficient search through the parameter space is hindered, leading to potentially slow and suboptimal outcomes and reliable feedback is often limited or absent, making it hard to determine the effectiveness of learning strategies.

In this section,  we provide the details of \textbf{BlackLaViP}, the gradient-free variant of LaViP. BlackLaViP utilizes \textit{Simultaneous Perturbation Stochastic Approximation} (SPSA) algorithm~\cite{ref_spsa1,ref_spsa2}.
SPSA approximate the gradient of high dimensional tensors by evaluating the difference between loss value under forward and backward perturbation. Given the positive decaying sequences of $a_i > 0$ and $c_i \in [0, 1]$, the gradient approximation, $\hat{g}$, and single-step parameter update of SPSA is described as follows:
\begin{align}
\hat{g}_i(\phi_{w_i}) = \frac{L(\phi_{w_i} + c_i \Delta_i) - L(\phi_{w_i} - c_i \Delta_i)}{2 c_i \Delta_i} \label{eq:gradient}
\end{align}
\begin{align}
\phi_{w_{i+1}} = \phi_{w_i} - a_i \hat{g}_i(\phi_{w_i}) \label{eq:update}
\end{align}
where $L$ is the loss function, $\phi_{w}\in\mathbb{R}^{d}$ is $d$-dimensional learnable parameters, and $\Delta_i \in \mathbb{R}^d$ is an $i$-th step random perturbation vector, sampled from mean-zero distributions.
Utilizing two consecutive forward evaluations, the SPSA interprets the estimated gradient through the difference in model output. This unique approach empowers us to optimize BlackLaViP's parameters $\phi_{w}$ without the reliance on backpropagation. 

\subsection{Implementation Details}
 We opt for $r=8$ and $r_{\mB}=32$ to effectively reduce the parameters, run all experiments for 300 epochs and evaluate the model on the last checkpoint. We have used parameters, perturbation distribution, and schedulers proposed by BlackVIP.

\begin{align*}
    a_i &= \frac{a}{(step_{i} + o)^{\alpha}} \\
    c_i &= \frac{c}{step_{i}^{\gamma}} \\
    total\_a_i\_steps &= Total\;Epochs \times {len(data\;loader)} \\
\end{align*}
Here, $a_i$ represents the learning rate and $c_i$ represents the perturbation scale, with specific values assigned as follows: 
\begin{align*}
    \alpha=0.4, a=0.01 , \gamma=0.2, o = \frac{{total\_a_i\_steps}}{\frac{\pi}{4}} \text{and } c=0.005\\
\end{align*}

\begin{figure*}[h!]
\centering
\includegraphics[width=0.85\textwidth]{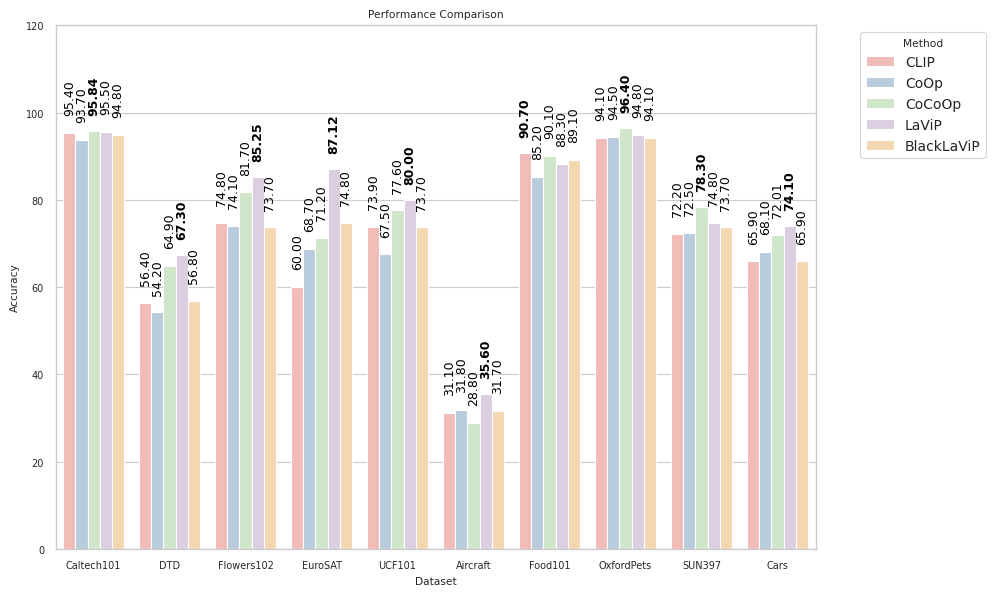} 
\caption{Performance of  BlackLaViP on base-to-novel generalization.Despite trailing behind other methods, BlackLaViP outperforms CoCoOp on EuroSAT and FGVCAircraft datasets. The best Harmonic Mean (HM) values on each dataset are bolded.}
\label{fig:bb_b2n}
\end{figure*}
\begin{table*}[t]
\centering
\resizebox{0.8\textwidth}{!}{
\begin{tabular}{l c c | l}
    \toprule
   Dataset & Classes & Parameters (M) & Prompt Template \\
   \midrule
   Caltech & 100 & 0.273 &\enquote{a photo of a \{\}.}\\
   Pets & 37 & 0.198 & \enquote{a photo of a \{\}, a breed of pet.}\\
   Cars & 196 & 0.403 &\enquote{a photo of a car model \{\}.}\\
   Flowers & 102 & 0.276 &\enquote{a vibrant photo of a \{\}, a type of flower.}\\
   Food & 101 & 0.275 &\enquote{a close-up photo showing delicious details of \{\}, a type of food.}\\
   Aircraft & 100 & 0.273 &\enquote{a photo of an aircraft \{\}.}\\
    SUN & 397 & 0.221 &\enquote{a photo showing good view of a \{\} location.}\\
   DTD & 47 & 0.210 & \enquote{a photo showing fine patterns of a \{\}-textured surface.}\\
   SVHN & 10 & 0.221 &\enquote{a pixelated street sign of the number \{\}.}\\
   EuroSAT & 10 & 0.221 &\enquote{a photo of centered satellite top-down view \{\} location.}\\
   RESISC & 45 & 0.207 &\enquote{a satellite imagery of a \{\} location.}\\
   CLEVR & 8 & 0.221 &\enquote{a photo of \{\} colored objects.}\\
   UCF & 101 & 0.275 &\enquote{a photo of a person actively engaged in \{\}.}\\
   CIFAR10 & 10 & 0.221 &\enquote{a pixelated photo of a \{\}.}\\
   CIFAR100 & 100 & 0.273 &\enquote{a pixelated photo of a\{\}.} \\
   \bottomrule
\end{tabular}

}
\caption{{Number of parameters and the text prompt template used in LaViP} }

\label{table:init_params}
\end{table*}

\subsection{Base-to-Novel Generalization}
Figure~\ref{fig:bb_b2n} compares BlackLaViP models in the context of generalization towards base-to-novel tasks. While it might fall short in comparison to other models in terms of performance, it is worth noting that BlackLaViP has displayed notable improvement over the CoCoOp model in the case of the EuroSAT and FGVCAircraft datasets, and exhibits marginally less on Caltech101 and OxfordPets. Moreover, it outperforms CoOp in 6 out of 10 datasets. We consider these outcomes to be strong indicators that the incorporation of language context holds the potential to mitigate the limitations observed in previous black-box models when it comes to generalizing to unseen classes.

 We would like to emphasize that the experiment was conducted with the sole intention of providing evidence for LaViP performance within a gradient-free environment. This demonstration holds substantial implications for future research, specifically in the realm of integrating language context into visual prompting for black-box training. While there might be a small loss of accuracy, the trade-off is more than justified by the advantage of achieving rapid convergence and sets a promising direction for exploring scenarios where faster convergence is of paramount importance with a small tradeoff of accuracy.

\section{Additional Ablation Studies}
\label{supp:ablation}
\subsection{Alternate Text Prompt Templates} We evaluate the effect of context words on the performance of LaViP. Table~\ref{tbl:fsl_template} presents the effect of utilizing different text prompt templates on LaVIP's performance. We employed two distinct prompt template types across the selected datasets.

\begin{itemize}
    \item \textit{{Base Prompt: }}Use same prompt template across all datasets: \enquote{\texttt{a\; photo\; of\; a\;\{\}}}
    \item {\textit{Custom Prompts:}} Building upon the base prompt template,  integrate context-specific visual tokens, as detailed in Table~\ref{table:init_params}

\end{itemize}
LaViP surpasses CLIP Zero-Shot prediction by a more significant margin when employing the base template. The inclusion of context-specific visual tokens further elevates performance. The results underscore the significance of language-grounded visual prompts in aligning vision-language modalities.

 \begin{table}
  \centering
  \resizebox{\columnwidth}{!}{
  \begin{tabular}{l| c c c c c c|c}
    \toprule
    Method & Caltech101 & DTD & Flowers102 & OxfordPets & EuroSAT &UCF101&\textit{Avg.}\\
    \midrule
    ZS & 89.3 & 46.0 & 70.4 & 88.9 & 54.1&69.8&69.75\\
    VP & 94.2 & 61.9 & 86.9 & 90.2 & 90.8 &74.2&83.03\\
    \midrule
    LaViP-Base & 95.0 & 67.7 & 96.2 & 90.3 & 86.0&81.1&86.05 \\
    
    \textbf{LaViP (Ours)}& \textbf{95.0} & \textbf{68.8} & \textbf{96.3 }& \textbf{91.2} & \textbf{86.2} &\textbf{81.3}&\textbf{86.47}\\
    \bottomrule
  \end{tabular}
  }
  \caption{Comparison of performance when using different text prompt templates} 
  \label{tbl:fsl_template}
\end{table}

\subsection{Language Integration Helps Domain Shift}
\begin{figure}[htp]
\centering
   
     \includegraphics[width=0.7\columnwidth]{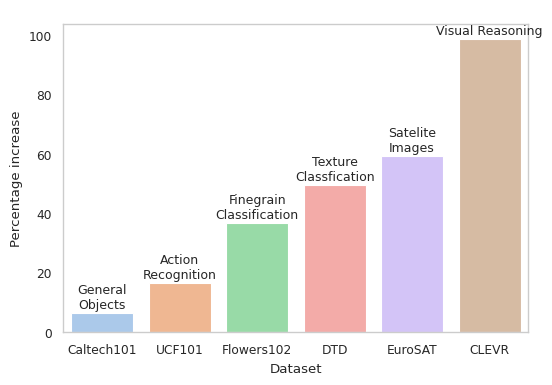}
   
\caption{Classification improvement of LaViP as domain shift increase($\rightarrow$) from pretraining dataset of CLIP} \label{fig:domain_shift}
\end{figure}

We evaluate LaViP on datasets characterized by less generic concepts and varying levels of perceptual diversity.  in Figure~\ref{fig:domain_shift}, we show the percentage of improvement achieved by LaViP over zero-shot CLIP. Notably, LaViP consistently outperforms when applied to datasets where domain shift from the pretraining dataset is more pronounced. 
LaViP surpasses CLIP Zero-Shot prediction by a more significant margin when employing the base template. The inclusion of context-specific visual tokens further elevates performance. The results underscore the significance of language-grounded visual prompts in aligning vision-language modalities.

\begin{figure}[h!]
\centering
\includegraphics[width=\columnwidth]{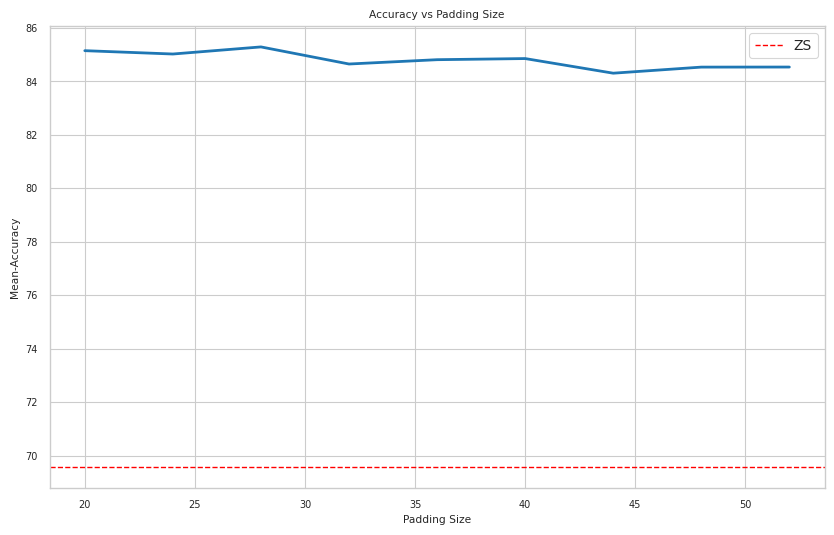} 
\caption{Average performance changes with different padding sizes. Using a moderate prompt size resulted in an overall improvement(LaViP uses $p=28$).LaViP outperforms CLIP zero-shot prediction by a significant margin on test padding sizes.}
\label{fig:fsl_padding}
\end{figure}
\subsection{Impact of Hyperparameters $(a, b,r)$} 

Table~\ref{tbl:r} presents the result obtained from 5 distinct datasets for varying $r$. LaViP robustly performs on these datasets when selecting $r$ such that $r \in [16,96]$
The effect of $a,b$ can be studied by varying the size of padding used for visual prompting.
 Figure \ref{fig:fsl_padding} illustrates the effect of padding size $p$ of LaViP. The average performance of LaViP exhibits fluctuations in response to different padding sizes. it is evident that our language-grounded visual prompts consistently outperform the zero-shot CLIP approach across all tested padding sizes. This clear trend highlights the efficacy of our model. Motivated by the ability to enhance overall performance across all datasets in our experiments, we selected a padding size of 28. Our decision also aligned with the observation previously reported by the VP.

 \begin{table}[t]

\footnotesize
    \centering
    \renewcommand{\arraystretch}{1}
    \renewcommand{\tabcolsep}{1.0mm}
    \resizebox{\columnwidth}{!}{
        \begin{tabular}{l|ccccccccc}

            \toprule
            
            Dataset & \textbf{1}& \textbf{2}& \textbf{4}& \textbf{8}& \textbf{16}& \textbf{32}& \textbf{64}& \textbf{96}&\textbf{LaViP (Ours)}\\
            \midrule
            Caltech101 &33.3&	33.5&	34.3&	34.7&	94.5	&94.9	&94.9	&94.9&95.0\\
        DTD & 31.1	&50.3	&49.1	&60.2	&64.6&	68.6	&68.8&	67.7&68.8\\ 
        EuroSAT& 77.2	&78.7	&79.9&	85.0&	85.7&	84.0&	86.1&	84.3 &86.1\\ 
        
        Flowers102 &44.7&	49.2&	26.6&	88.8	&93.8&	95.2&	96.2&	96.1&96.3\\ 
        OxfordPets&89.6	&90.4	&90.8	&90.9&	90.5&	91.2&	90.8&	90.8&91.2\\
         
        \midrule
        \textit{Average}&55.17	&60.42	&56.15&	71.90	&85.83	&86.77	&87.36	&86.77&87.48\\
        \bottomrule
        \end{tabular}
    }
    \caption{Performance of LaViP on 5 distinct datasets with varying $r.$}
    \label{tbl:r}
\end{table}

\subsection{Limitations of LaviP} 

\begin{table}
  \centering
  \resizebox{0.7\columnwidth}{!}{
  \begin{tabular}{l|c c c c }
    \toprule
    Method & SVHN & CIFAR10 & CIFAR100&PCam\\
    \midrule
    ZS &35.3 &\textbf{91.6}&\textbf{68.7}&48.1\\
    VP & \textbf{60.4}&73.2&49.9&\textbf{73.1}\\
    \midrule
    \textbf{LaViP (ours)}& 58.9&75.6&43.8&68.2\\
    \bottomrule
  \end{tabular}
    }
  \caption{Classification accuracy on datasets with images having low-resolution, and less semantic variation. LaViP falls short on datasets that have (a) low resolution (b) more generic concepts (c) less semantic variation. } 
  \label{tbl:low_res}
\end{table}
Table \ref{tbl:low_res} shows the performance of LaViP on low-resolution datasets compared to CLIP and VP. We hypothesize that the observed characteristic of LaViP can be attributed to the challenge of context tokens failing to establish meaningful interactions with the semantic content of the images due to information loss when upscaling them(SVHN, CIFAR10, CIFAR100). In cases where concept diversity is elevated (such as CIFAR10 and CIFAR100), the utilization of a single prompt template proves less effective in accommodating the wide range of contexts inherent to these concepts. Moreover, when the semantic variability of images(PCam) is diminished, the language template fails to offer meaningful context information that could facilitate improved interaction.
\begin{table}[t]
\footnotesize
    \centering
    \renewcommand{\arraystretch}{1}
    \renewcommand{\tabcolsep}{1.0mm}
    \resizebox{\columnwidth}{!}
    {

    \begin{tabular}{lccccccccccccccc}
        \toprule
        
       \multicolumn{3}{c}{\textbf{Source}} & \multicolumn{10}{c}{\textbf{Target}} \\
       & \rotatebox{60}{ImageNet} & \rotatebox{60}{Caltech101} & \rotatebox{60}{OxfordPets} & \rotatebox{60}{StanfordCars}&\rotatebox{60}{Flowers102} & \rotatebox{60}{Food101}& \rotatebox{60}{FGVCAircraft} &\rotatebox{60}{SUN397} & \rotatebox{60}{DTD} & \rotatebox{60}{EuroSAT} & \rotatebox{60}{UCF101} & \rotatebox{60}{Average} \\
       \midrule
        CoOp &  \textbf{71.51} & 93.70 & 89.14 & 64.51 & 68.71 & 85.30 & 18.47 & 64.15 & 41.92 & 46.39 & 66.55 & 63.88\\
        Co-CoOp & 71.02 & \textbf{94.43} & 90.14 & 65.32 & 71.88 & 86.06 & 22.94 & \textbf{67.36}&45.73&45.37 & 68.21 & 65.74 \\
        MaPLe & 70.72 & 93.53 & \textbf{90.49} & \textbf{65.57} & \textbf{72.23} & \textbf{86.20} & 24.74 & 67.01 & \textbf{46.49} & 48.06 & \textbf{68.69} & 66.30 \\
        
        \midrule
        \textbf{LaViP (Ours)} & 65.97 & 93.10& 89.63 & 64.87 & 70.53&85.73& \textbf{25.23}&63.47&46.30&\textbf{58.87}&68.17&\textbf{66.53}& \\
   
        \bottomrule
    \end{tabular}
    }
    \caption{Comparison of LaViP with previous methods on cross-dataset evaluation.}

\label{tbl:xd}
\end{table}

\subsection{LaViP adaptability to CLIP variants}

We evaluate to assess the compatibility of LaViP with the CNN and ViT backbones, available in CLIP. Table~\ref{tbl:clip_var} presents the outcomes for the DTD dataset, wherein a learning rate of 0.01 was used for RN50 and RN101. The table reveals that LaViP surpasses 2 out of 3 of the tested variants, including the CNN backbone.
Compared to VP, LaViP achieves an absolute gain of \textbf{8.5\%} over CLIP zero shot with an optimization that is \textit{more than} $3\times$ faster. This result highlights that LaViP can seamlessly adapt to diverse backbone architectures while maintaining or even improving performance.

\begin{table}[h]
\footnotesize
    \centering
    \renewcommand{\arraystretch}{1}
    \renewcommand{\tabcolsep}{1.0mm}
    \resizebox{\columnwidth}{!}
    {

    \begin{tabular}{lccccc}
        \toprule
        
      \multicolumn{3}{c}{\textbf{Source}} & \multicolumn{3}{c}{\textbf{Target}} \\
       &ImageNet &ImageNetV2 &Imagenet-S&ImageNet-A &ImageNet-R\\
       \midrule
        CLIP & 66.73 &60.83 &46.15 &47.77 & 73.96\\
        CoOp &  \textbf{71.51} & \textbf{64.20}&47.99&49.71&75.21\\
        Co-CoOp &  71.02&64.07&48.75&50.63&76.18\\
        MaPLe & 70.72 &64.07 & \textbf{49.15} & \textbf{50.90} & 76.98\\
        \midrule
        \textbf{LaViP (Ours)} & 65.95 &61.60& 47.23& 48.91 &\textbf{83.93}\\
   
        \bottomrule
    \end{tabular}
    }
    \caption{Comparison of LaViP with existing approaches in the domain generalization setting.}

\label{tbl:dg}
\end{table}
\begin{table}[t]
  \centering
  \resizebox{0.7\columnwidth}{!}{
  \begin{tabular}{l|c c c|c}
    \toprule
    Method & RN50 & RN101 &ViT-B/16&\textit{Avg.}\\
    \midrule
    ZS &41.7&43.9&46.0&43.87\\
    VP & \textbf{46.0}&33.3&61.9&47.07\\
    \midrule
    \textbf{LaViP (Ours)}& 42.2&\textbf{46.1}&\textbf{68.8}&\textbf{52.37}\\
    \bottomrule
  \end{tabular}
    }
  \caption{Experiment study for backbone variant of CLIP on the few-shot learning task. Classification accuracy on DTD across pre-trained backbone architectures of CLIP including the CNN and ViT.} 
  \label{tbl:clip_var}
\end{table}

\subsection{Cross-dataset Generalization}

We test the cross-dataset generalization ability of LaViP by learning visual prompts on all the 1000 Imagenet~\cite{ref_imagenet} classes and transferring it to the remaining datasets using our Kronecker product knowledge fusion.
Table~\ref{tbl:xd} shows the performance comparison between LaVIP, CoOp, CoCoOp and MaPLe. On the ImageNet, the performance of LaViP is suboptimal. We hypothesize this is due to the domain of ImageNet having a strong correlation with the pretraining dataset of CLIP. We used NVIDIA A100 GPU for this experiment with batch size 64 and prompt template \enquote{\texttt{a cropped photo of a < class >}}

\subsection{Domain Generalization}
To understand the domain generalization of LaViP, we transfer the ImageNet mode to various out-of-domain (OOD) datasets. Table\ref{tbl:dg} shows the comparison of domain generalization capability of LaViP compared to CLIP, CoOp, CoCoOp and MaPLe. The performance of the ImagNet dataset impacts the performance of OOD datasets.

\end{document}